# Organization and Understanding of a Tactile Information Dataset TacAct During Physical Human-Robot Interactions

Peng Wang, Jixiao Liu, *Member, IEEE*, Funing Hou, Dicai Chen, Zihou Xia, and Shijie Guo, *Member, IEEE*

*Abstract*— Advanced service robots require superior tactile intelligence to guarantee human-contact safety and to provide essential supplements to visual and auditory information for human-robot interaction, especially when a robot is in physical contact with a human. Tactile intelligence is an essential capability of perception and recognition from tactile information, based on the learning from a large amount of tactile data and the understanding of the physical meaning behind the data. This report introduces a recently collected and organized dataset "TacAct" that encloses real-time pressure distribution when a human subject touches the arms of a nursing-care robot. The dataset consists of information from 50 subjects who performed a total of 24,000 touch actions. Furthermore, the details of the dataset are described, the data are preliminarily analyzed, and the validity of the collected information is tested through a convolutional neural network LeNet-5 classifying different types of touch actions. We believe that the TacAct dataset would be more than beneficial for the community of human interactive robots to understand the tactile profile under various circumstances.

## I. INTRODUCTION

Natural interaction through touch is one of the ultimate goals for many scenarios involving physical Human-Robot Interactions (pHRI) [1]. Touching can directly measure various properties of the contact objects and the external environment, and it is one of the indispensable ways for an intelligent robot to perceive the environment [2]. In most pHRI scenarios, physical touch plays a central role in various applications of service robots, since nonverbal interaction behavior is a key factor that makes the process more natural [3]. An essential aspect of human-robot interaction is to respond to different touch stimuli, which usually needs a robot equipped with tactile sensors and a touch action recognition system [4]. The algorithm of the action recognition system generally requires a dataset for training and recognition.

In recent years, there has been some related research work on the human gestures dataset. The Corpus of Social Touch [5], CoST, was collected using the pressure sensor wrapped around the human arm model, which contains 14 different social touch gestures of 7805 data samples, and the final recognition accuracy can reach 60% using various classifiers [6]; the HAART dataset [7] was collected using a fabric sensor, which was wrapped in the skeleton of the CuddleBot cat-shaped robot. The classification includes 5 different touch gestures, namely continuous slap, pat, scratch, stroke, and tickle. The accuracy of the random forest model recognition reaches 90%. From the analysis and recognition results of these two datasets, it can be seen that the recognition accuracy is not enough when there are more types of actions, while fewer types of actions cannot express the diversity of human-robot interaction. Therefore, a pHRI dataset with larger-scale and more classifications is needed.

In this paper, we introduce a dataset "TacAct" containing tactile information from 12 common human touching actions of 50 subjects. Analysis suggests rich information contained from the multi-dimensional tactile data. The purpose of collecting and organizing the tactile information dataset is to describe the processes of human touching robots, to understand the pHRI, and to further build algorithms for recognizing human intentions.

## II. THE "TACACT" DATASET

To address the need for physical Human-Robot Interactions actions dataset that can express human intentions, we recorded 12 common tactile actions of 50 subjects with a tactile sensor. The "TacAct" dataset is publicly available[1].

### A. The Tactile Sensor

Collecting a dataset generally requires large-area, high-resolution flexible array tactile sensors, and the sensor sampling frequency and pressure measurement range need to meet the requirements. The tactile sensor in this paper was

Peng Wang is with the school of Mechanical Engineering, Hebei University of Technology, China and Hebei Key Laboratory of Smart Sensing and Human-Robot Interaction, Hebei University of Technology, Tianjin 300132 (e-mail: 201831204052@stu.hebut.edu.cn).

Jixiao Liu is with the school of Mechanical Engineering, Hebei University of Technology, China and Hebei Key Laboratory of Smart Sensing and Human-Robot Interaction, Hebei University of Technology, Tianjin 300132 (corresponding author, phone: 86-185-2209-9429; e-mail: liujixiao@hebut.edu.cn).

Funing Hou is with the school of Academy for Engineering&Technology, Fudan University, China and Hebei Key Laboratory of Smart Sensing and Human-Robot Interaction, Hebei University of Technology, Tianjin 300132.

Dicai Chen and Zihou Xia are with the school of Mechanical Engineering, Hebei University of Technology, China and Hebei Key Laboratory of Smart Sensing and Human-Robot Interaction, Hebei University of Technology, Tianjin 300132.

Shijie Guo is with the school of Mechanical Engineering, Hebei University of Technology, China and Hebei Key Laboratory of Smart Sensing and Human-Robot Interaction, Hebei University of Technology, Tianjin 300132.

[1]Data available on request.

improved based on the previous study [8] and the pressure spatial distribution can be obtained when using the tactile sensor. The tactile sensor consists of three layers of materials: two layers of membranes of screen-printed flexible electrodes and a middle layer of ionic-gel microfiber matrix [9].

The principle of the tactile sensor is shown in Figure 1(a). When pressure is applied, the ionic-gel microfiber layer is compressed, the contact area between the middle layer and the upper and lower electrodes increases, and the contact nodes in the fiber increase, resulting in a change in capacitance ($C$ to $C'$) and resistance ($R$ to $R'$).

The sensor array is 32×32, the electrode width is 5 mm, the moment between adjacent electrodes is 1 mm, the size of the applied sensing area is 191 × 191 mm$^2$ and about 0.5 mm thick, as shown in Figure 1(b). The upper electrode layer of the sensor consists of 32 column electrodes, and the lower electrode layer consists of 32 row electrodes. The sensor area is gridded in 1024 elements, and one gridded element size is 25 mm$^2$. The pressure sensor grid detectable pressure ranges from 2×10$^{-5}$ to > 0.2 MPa at an ambient temperature of 25 °C. The size and the measuring range of this sensor are suitable for measuring the touching behaviors of human hands.

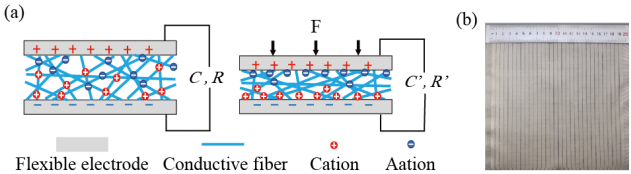

Figure 1. The flexible supercapacitor tactile sensor. (a) Working principle of the tactile sensor. (b) Picture of the tactile sensor.

*B. Measurement Principle of Tactile Sensor*

The capacitive reactance at the intersection of the row electrode and the column electrode of the array tactile sensor can be expressed as

$$X_c = \frac{1}{2\pi fC} \quad (1)$$

where $X_c$ is the capacitive reactance; $f$ is the frequency of the sinusoidal signal source; $C$ is the capacitance of a sensing unit.

As shown in Figure 2, the upper electrode of the sensor is connected to the sinusoidal signal source, the lower electrode is connected to the sampling capacitor $C_0$, the sensing unit is connected in series with the sampling capacitor to form a voltage divider circuit.

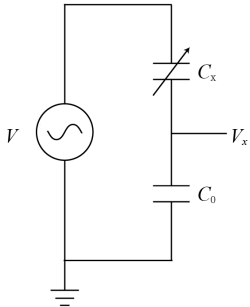

Figure 2. Principle of pressure index acquisition circuit

The voltage of the sampling capacitor can be expressed by

$$V_x = \frac{1}{\frac{C_0}{C_x}+1} \cdot V \quad (2)$$

where $V_x$ is the sampling voltage, $V$ is the sinusoidal voltage source, $C_0$ is the sampling capacitance, and $C_x$ is the capacitance of the current measurement point. Therefore, the magnitude of the force can be characterized by the voltage signal.

The pressure index (unit-less) stored by the computer is the proportional value of the collected voltage and does not indicate the exact pressure. There is a linear relationship between pressure and capacitance, and the voltage detected by the circuit is positively correlated with pressure, so the pressure index can be used to characterize the pressure.

*C. Physical Human-Robot Interaction Actions*

We selected 20 common different physical human-robot contact actions from the touch studied by Yohanan [3] and designed a questionnaire, which was distributed to 80 researchers in the robotics field. The questionnaire asked to select 12 out of 20 common actions that are most suitable for human-robot interaction and collaboration. Finally, 57 valid questionnaires were received, and the actions selected were shown in Table I.

TABLE I. LIST OF HUMAN ACTIONS IN THE DATASET

| Actions | Description |
|---|---|
| #1 Pull | Pull inward to the subjects with fingers/palm |
| #2 Squeeze | With two or three fingers |
| #3 Push | Push outward from the subjects with palm |
| #4 Hold | With entire hand (including palm) |
| #5 Grasp | With five fingers (without palm) |
| #6 Poke | With one finger |
| #7 Static drag | With entire hand dragging |
| #8 Strongly hit | Rapidly hit with force (once) |
| #9 Soft slide | Touch and move with hands |
| #10 Scratch | With fingers moving |
| #11 Soft tap | Rapidly and softly tap (twice) |
| #12 Sliding drag | With entire hand drag and move |

*D. Data Collection*

The dataset was collected by a flexible supercapacitor tactile sensor installed in an imitated mechanical arm device. In a 32 × 32 grid, the sensor data can be sampled at 100 Hz (100 frames per second). A total of 12 touch actions, namely, pull, squeeze, push, hold, grasp, poke, static drag, strong hit, soft slide, scratch, soft tap, and sliding drag, were recorded, as shown in Figure 3. A single-action collected from a subject consists of a 32×32×$N$ matrix (where $N$ is the number of frames or frame length). For the same action, the subjects were asked to use as many postures as possible to apply different magnitudes of force on different positions of the sensor. All subjects were also asked to perform all actions in right and left hand, repeated 20 times with each hand. Except for strongly hit and soft tap (complete in an instant, acquisition time 2s), the duration of each action is 2s, and the time is 4s altogether. The experiment was conducted on 50 subjects (36 males and 14 females，ranged from 22 to 36 years and 44 were right-handed) in total, each subject consists of 480 actions (12 actions × 40 repetitions ) in total, and the dataset collected 24,000 actions from the subjects.

The collected 20 consecutive repetitive actions are divided into 20 samples, and the tap and tap are divided into 2s, that is, the frame length of instantaneous action is 200; other actions are divided into 4s, the frame length of continuous action is 400. The number of repetitions of the same action is 40 times, and the number of each action in the dataset is 2,000. The dataset has a total of 24,000 training samples, each frame in each training sample contains 1,024 pressure index values, corresponding to the pressure at the intersection of each row and column of the 32×32 array sensor pressure index value.

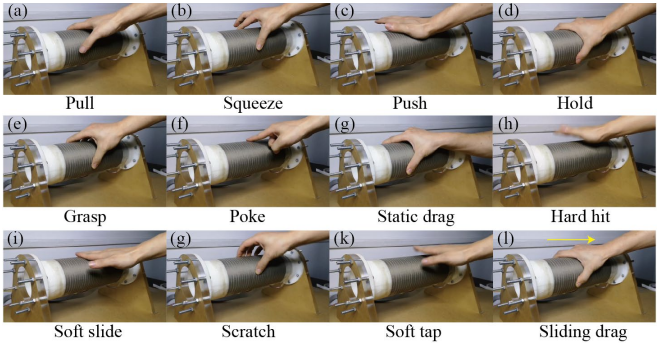

Figure 3. Demonstration of 12 actions.

## III. PHYSICAL FEATURES EXTRACTION

To understand pHRI from the TacACT dataset, this article lists four contact action features and discussed the correlation of actions between the subjects.

### A. Touching Behaviors and Key Frame

Taking the touching behavior of 'hold' as an example, pressure index data were sampled at a rate of 100 Hz. The total number of frame sequences is 400 (100×4). A typical 'hold' mode sequence is shown in Figure 4. The hold action sequence consists of 10 frames (= 100 msec). As shown in this figure, the sampling rate is fast enough to capture the sharp pressure changes in the hold action.

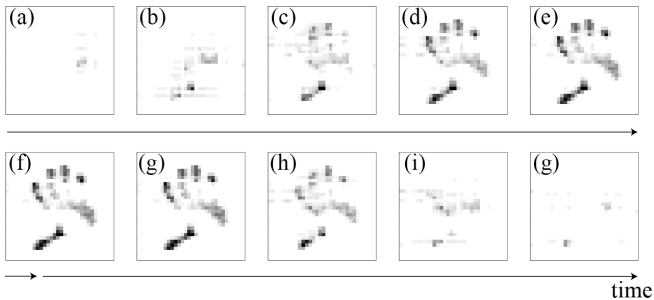

Figure 4. A typical 'hold' sequence (1 frame = 1/100 sec = 10 msec). (a)~(g) describe the process of the 'hold' action

The sum of the pressure matrix of each frame is calculated, and the frame with the largest pressure matrix and value is selected as the keyframe of each action sample. We convert the keyframes (maximum mean data frame in an action) of tactile pressure matrices into a grayscale image to visually illustrate the physical contact. Figure 5 lists 36 selected actions (three picked for each action randomly). Intuitively, each action has its obvious characteristics, which provides the possibility for action recognition to express intent.

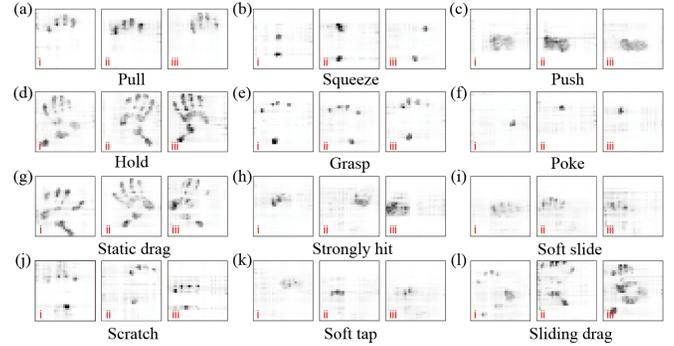

Figure 5. Spatial distribution of keyframes of contact pressure for all 12 kinds of actions.

### B. Features Extraction

As the first point of understanding human-robot interaction, we use the following features as the basic characteristics of pHRI actions. The feature calculation formula is as follows:

- $\omega(t)$: The mean value of pressure index for a frame at time $t$

$$\omega(t) = \frac{1}{1024}\sum_i\sum_j F_{ij}(t) \quad (1 \leq i, j \leq 32) \quad (3)$$

- $v(t)$: The change speed of pressure index for a frame at time $t$

$$v(t) = \omega(t) - \omega(t - \Delta t) \quad (t \geq \Delta t) \quad (4)$$

- $a(t)$: The total contact area of a frame at time $t$

$$a(t) = \sum_i\sum_j b_{ij}(t) \quad (1 \leq i, j \leq 32) \quad (5)$$

$$b_{ij}(t) = \begin{cases} 1, & F_{ij}(t) \geq f_{\text{thresh}} \\ 0, & \text{otherwise} \end{cases} \quad (6)$$

- $(x_t, y_t)$: The position of the centroid at time $t$

$$P_{00} = \sum_i\sum_j F_{ij}(t) \quad (1 \leq i, j \leq 32) \quad (7)$$

$$P_{10} = \sum_i\sum_j i \times F_{ij}(t) \quad (1 \leq i, j \leq 32) \quad (8)$$

$$P_{01} = \sum_i\sum_j j \times F_{ij}(t) \quad (1 \leq i, j \leq 32) \quad (9)$$

$$x_t = \frac{P_{10}}{P_{00}} \quad (10)$$

$$y_t = \frac{P_{01}}{P_{00}} \quad (11)$$

where $F_{ij}(t)$ is the pressure value in the i-row and j-column of the tactile sensor.

The four features applied by the right hands of subject number 24 (female) were randomly selected to analyze the dataset. As shown in Figure 6, there is a big gap in the characteristics of different actions.

## C. Relevance of Actions

Figure 7 and Figure 8 show the evolution over frame of the summed pressure for the action 'hold' performed by a single subject and multiple different subjects, respectively. These examples illustrate the variation in duration and intensity of the action, within a subject and between subjects.

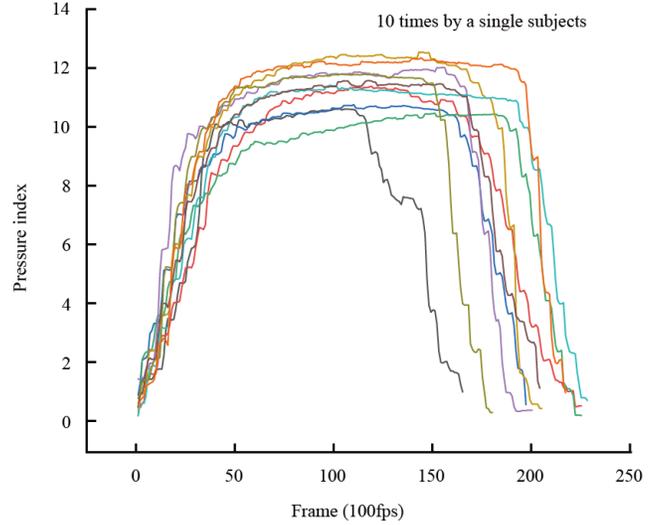

Figure 7. Meaned pressure index per frame of 'hold' performed ten times by a single subject

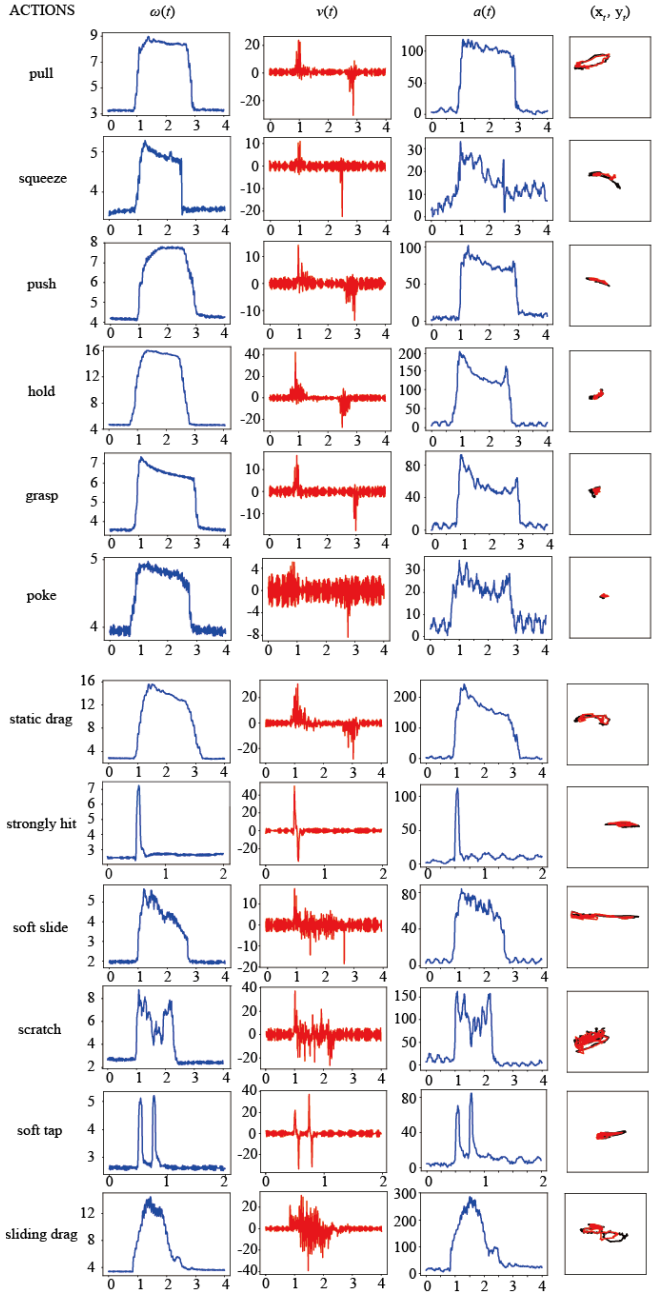

Figure 6. Features of different actions

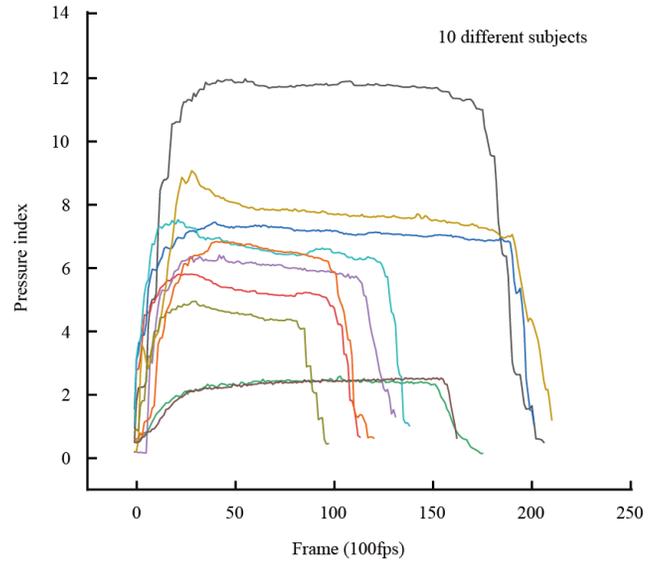

Figure 8. Meaned pressure index per frame of 'hold' performed by ten different subjects

For the same kind of actions, the difference in pressure amplitude and action duration between the same subject is very small, however, the gap between different subjects is very large.

## IV. CLASSIFICATION

As one of the representative algorithms of deep learning, convolutional neural networks are widely used in image recognition problems. Since the pressure index matrix of each frame of the action can be regarded as a two-dimensional gray image, CNN can be used for classification [10]. To verify the validity of the TacAct dataset, a simple and common CNN structure is used for classification. We explored a convolutional neural network architecture, LeNet-5, to classify actions. It can use frames for classification within a short time.

### A. Data Preprocessing

The initial frame of no action is subtracted from each frame of each action sample to eliminate noise interference. The pressure index threshold is set to 20. When any point in the pressure index matrix is bigger than the pressure index threshold, it is considered that the human and the robot are in contact with each other and the action starts; in other cases, the contact action is considered to be completed. Based on this, the effective action process in the action sample is extracted, which provides convenience for the subsequent information mining and the calculation of the centroid change in the action process.

### B. LeNet-5

Based on the LeNet-5 network, a convolutional neural network suitable for tactile matrix input was established to verify the validity of the dataset. The CNN structure including an input layer, three convolutional layers, two pooling layers, a fully connected layer, and an output layer, as shown in Figure 9. The input of tactile information is a tensor of $32\times32\times N$, where $32\times32$ represents the data information of each frame, and $N$ denotes the length of the frame. If the number of frames is insufficient, use null matrix completion. This paper tests the accuracy of actions with various frame numbers ($N$) from 5 consecutive frames to 100 frames, and the results are shown in Figure 10.

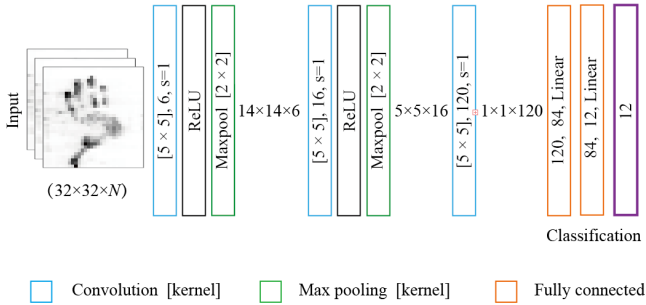

Figure 9. LeNet-5 Architecture.

## V. CONCLUSION

It can be seen that the data of TacAct is diverse from the analysis of the action features. This proves that the dataset has strong generalization ability. The results also suggest that if frame length (N) is within a certain range, the increase of N will raise the accuracy of action recognition. After N exceeds a certain value, the accuracy hardly continues to improve. When the frame length reaches 30, the recognition accuracy rate exceeds 90%; the highest recognition accuracy rate is 95.46% when the frame length is 80. The increase in $N$ leads to an increase in input channels, thereby adding the convolution operation, which is computationally expensive.

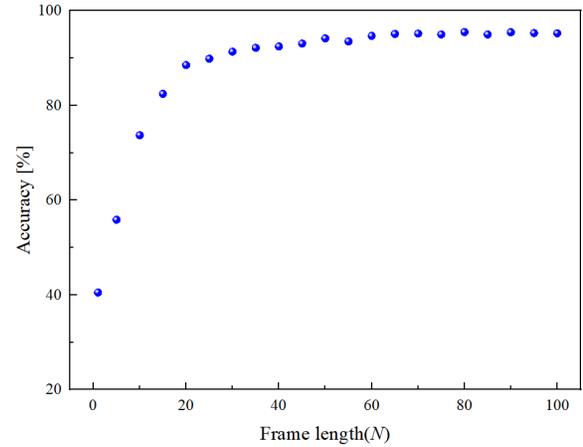

Figure 10. Accuracy of actions with varies frame length ($N$) from 5 consecutive frames to 100 frames

Therefore, it is necessary to comprehensively consider accuracy and calculation cost when choosing the network. The LeNet-5 reached a high accuracy which proves the effectiveness of the dataset.

REFERENCES

[1] Q. Li, O. Kroemer, Z. Su, F. F. Veiga, M. Kaboli and H. J. Ritter, "A Review of Tactile Information: Perception and Action Through Touch," in *IEEE Transactions on Robotics*, vol. 36, no. 6, pp. 1619-1634, Dec. 2020, doi: 10.1109/TRO.2020.3003230.

[2] A. Albini, S. Denei and G. Cannata, "Human hand recognition from robotic skin measurements in human-robot physical interactions," *2017 IEEE/RSJ International Conference on Intelligent Robots and Systems (IROS)*, Vancouver, BC, Canada, 2017, pp. 4348-4353, doi: 10.1109/IROS.2017.8206300.

[3] Yohanan, Steve, and K. E. Maclean. "The Role of Affective Touch in Human-Robot Interaction: Human Intent and Expectations in Touching the Haptic Creature." *International Journal of Social Robotics* 4.2(2012):163-180.

[4] F. Naya, J. Yamato and K. Shinozawa, "Recognizing human touching behaviors using a haptic interface for a pet-robot," *IEEE SMC'99 Conference Proceedings. 1999 IEEE International Conference on Systems, Man, and Cybernetics (Cat. No.99CH37028)*, Tokyo, Japan, 1999, pp. 1030-1034 vol.2, doi: 10.1109/ICSMC.1999.825404.

[5] Jung, Merel M., et al. "Touching the Void -- Introducing CoST." *International Conference* 2014:120-127.

[6] Jung, Merel M., et al. "Automatic recognition of touch gestures in the corpus of social touch." *Journal on Multimodal User Interfaces* (2016).

[7] Cang, Xi Laura, et al. "Different Strokes and Different Folks: Economical Dynamic Surface Sensing and Affect-Related Touch Recognition." *the 2015 ACM* ACM, 2015.

[8] J. Liu *et al.*, "Cost-Efficient Flexible Supercapacitive Tactile Sensor with Superior Sensitivity and High Spatial Resolution for Human-Robot Interaction," in *IEEE Access*, vol. 8, pp. 64836-64845, 2020, doi: 10.1109/ACCESS.2020.2984511.

[9] Bai, Ningning, et al. "Graded intrafillable architecture-based iontronic pressure sensor with ultra-broad-range high sensitivity." *Nature Communications* (2020).

[10] H. Wu, D. Jiang and H. Gao, "Tactile motion recognition with convolutional neural networks," *2017 IEEE/RSJ International Conference on Intelligent Robots and Systems (IROS)*, Vancouver, BC, Canada, 2017, pp. 1572-1577, doi: 10.1109/IROS.2017.8205964.